%% file: VizExtract.tex
\documentclass[sigconf, nonacm]{acmart}

\newcommand\vldbdoi{XX.XX/XXX.XX}
\newcommand\vldbpages{XXX-XXX}
\newcommand\vldbvolume{14}
\newcommand\vldbissue{1}
\newcommand\vldbyear{2021}
\newcommand\vldbauthors{\authors}
\newcommand\vldbtitle{\shorttitle} 
\newcommand\vldbavailabilityurl{github.com/ddecatur/VizExtract}
\newcommand\vldbpagestyle{plain} 

\usepackage{graphicx}
\usepackage{textcomp}
\usepackage{xcolor}
\usepackage{xspace}
\usepackage{pdfrender}
\usepackage{algpseudocode}
\usepackage{enumitem}
\usepackage{algorithm}
\usepackage{appendix}

\newcommand*{\boldcheckmark}{%
  \textpdfrender{
    TextRenderingMode=FillStroke,
    LineWidth=.5pt, 
  }{\checkmark}%
}
\newcommand\sys{VizExtract\xspace}

\begin{document}
\title{\sys: Automatic Relation Extraction from Data Visualizations}

\author{Dale Decatur}
\affiliation{%
  \institution{University of Chicago}
  \city{Chicago}
  \state{Illinois}
}
\email{ddecatur@uchicago.edu}

\author{Sanjay Krishnan}
\affiliation{%
  \institution{University of Chicago}
  \city{Chicago}
  \state{Illinois}
}
\email{skr@uchicago.edu}

\begin{abstract}
Visual graphics, such as plots, charts, and figures, are widely used to communicate statistical conclusions.
Extracting information directly from such visualizations is a key sub-problem for effective search through scientific corpora, fact-checking, and data extraction.
This paper presents a framework for automatically extracting compared variables from statistical charts. 
Due to the diversity and variation of charting styles, libraries, and tools, we leverage a computer vision based framework to automatically identify and localize visualization facets in line graphs, scatter plots or bar graphs and can include multiple series per graph. 
The framework is trained on a large synthetically generated corpus of \texttt{matplotlib} charts and we evaluate the trained model on other chart datasets. 
In controlled experiments, our framework is able to classify, with $87.5\%$ accuracy, 
the correlation between variables for graphs with 1-3 series per graph, varying colors, and solid line styles. When deployed on real-world graphs scraped from the internet, it achieves $72.8\%$ accuracy ($81.2\%$ accuracy when excluding ``hard" graphs). When deployed on the FigureQA dataset, it achieves $84.7\%$ accuracy.
\end{abstract}

\maketitle

\pagestyle{\vldbpagestyle}
\begingroup\small\noindent\raggedright\textbf{PVLDB Reference Format:}\\
\vldbauthors. \vldbtitle. PVLDB, \vldbvolume(\vldbissue): \vldbpages, \vldbyear.\\
\href{https://doi.org/\vldbdoi}{doi:\vldbdoi}
\endgroup
\begingroup
\renewcommand\thefootnote{}\footnote{\noindent
This work is licensed under the Creative Commons BY-NC-ND 4.0 International License. Visit \url{https://creativecommons.org/licenses/by-nc-nd/4.0/} to view a copy of this license. For any use beyond those covered by this license, obtain permission by emailing \href{mailto:info@vldb.org}{info@vldb.org}. Copyright is held by the owner/author(s). Publication rights licensed to the VLDB Endowment. \\
\raggedright Proceedings of the VLDB Endowment, Vol. \vldbvolume, No. \vldbissue\ %
ISSN 2150-8097. \\
\href{https://doi.org/\vldbdoi}{doi:\vldbdoi} \\
}\addtocounter{footnote}{-1}\endgroup

\ifdefempty{\vldbavailabilityurl}{}{
\vspace{.3cm}
\begingroup\small\noindent\raggedright\textbf{PVLDB Artifact Availability:}\\
The source code, data, and/or other artifacts have been made available at \url{\vldbavailabilityurl}. 
\endgroup
}

\input{Introduction}

\input{RelatedWork}
\input{SystemArchitecture}
\input{Prototype}

\input{Evaluation}
\input{Conclusion}


\bibliographystyle{ACM-Reference-Format}
\bibliography{Bib}

\clearpage
\appendix

\input{app-segment}

\end{document}

%% file: Introduction.tex
\section{Introduction}
Organizing and indexing data products, such as reports, charts, and models, is a core aspect of data governance~\citep{perkel2018jupyter, hellerstein2017ground}.
Data scientists need to be able to quickly identify how datasets have been used, by whom, and what the conclusions were.
Generally, this process involves extracting metadata from analysis code and any derived data artifacts and storing that metadata in a database~\citep{hellerstein2017ground, fernandez2018aurum}.
The extraction problem has been extensively studied in the context of tabular analytics~\citep{fernandez2018aurum, nargesian2019data, rehmandemonstration}, raw text (e.g., publications)~\citep{niu2012deepdive}, and machine learning~\citep{zaharia2018accelerating, garcia2018context}. 

Notably missing from this body of work are reliable tools that can extract information from data visualizations.
Several research prototypes have been proposed: Scatteract, FigureQA, ChartText, ChartOCR, and Reverse Engineering Visualizations (REV) ~\citep{luo2021chartOCR, cliche2017scatteract, poco2017reverse, kahou2018figureQA, balaji2018dcharttext}. 
However, our literature survey suggests relatively narrow applications for these systems focusing on certain graph types, styles, and facets.
For example, REV focuses on extracting textual fields such as legends, axis labels, and ticks from a visualization, ignoring the actual data ~\citep{poco2017reverse}.
On the other hand, ChartOCR ~\citep{luo2021chartOCR} attempts to extract all information, including the data, but that can result in poor performance on ``charts in the wild'' due to unanticipated charting styles. 
Accordingly, this paper explores the methodology and limits of a general-purpose visualization information extraction system. 
Fundamentally, this problem is a computer vision problem, in which an AI system needs to be able to segment and classify different facets of an image file representing the visualization.
We show that with an appropriately defined extraction scope, a deep learning system can effectively identify key visualization aspects across varied chart types, styles, and arrangements.

\begin{figure}
    \centering
    \includegraphics[width=0.8\columnwidth]{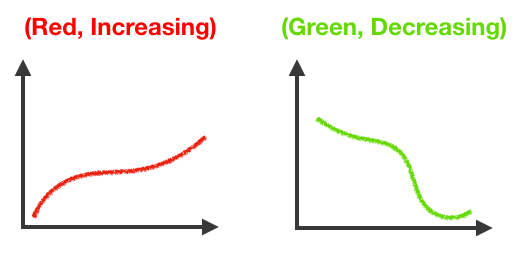}
    \caption{Examples of visual relations. Visual relations describe how a particular chart series relates to the X-Y plane.}
    \label{fig:rels} \vspace{-1em}
\end{figure}

This paper proposes \sys, a visualization information extraction system.
\sys extracts axis labels, legends, and series from bar graphs, scatter plots, and line graphs.
In terms of plot content, \sys focuses on identifying high-level relationships among the visualized data rather than identifying every individual mark.
This is similar in spirit to relation extraction in natural language processing, where the objective is to identify binary relations between entities.
Similarly, \textbf{visual relations} are binary relationships between the variables described on a chart (i.e., the variables represented on the x and y axes) conditioned on a particular chart series. 
For example, one could tag every series with a ``correlation'' (positive, negative, or neutral). 
This correlation would describe if the series is generally increasing on Y w.r.t X, decreasing, or constant (Figure \ref{fig:rels}).
Thus, in visual relation extraction, given an input chart represented as an image, we would like to yield 4-tuples of the form (\textsf{x-axis}, \textsf{y-axis}, \textsf{series}, \textsf{relationship}). For the purposes of this paper, we scope visual relations to solely determining correlation, but this same approach should hold for any relation.

In practice, the implementation of such a system is quite complex because charts have varying styles, multiple series, and are generated using a diverse range of graphing libraries. We contribute a computer vision pipeline, which given an input chart as an image, extracts a set of discovered relations. The framework is based on a deep learning model that is trained on a large synthetically generated corpus of \texttt{matplotlib} charts. These charts are varied in their organization, style, and coloring.
The trained model can then be deployed on unseen charts and easily fine-tuned for new styles.
In developing \sys, we have learned much about the strengths and weaknesses of deep learning in this context. While deep learning is an immensely powerful tool and is the foundation of our system, our research suggests that the approach of using solely a deep learning network is not sufficient to achieve accurate results and a mix of learned and programmed components is necessary to achieve best performance. \sys uses a combination of pre-processing and other algorithmic approaches surrounding its central convolutional neural network (CNN) in order to deliver its results.

We evaluated \sys against a state-of-the-art baseline called ChartOCR ~\citep{luo2021chartOCR}. We chose to compare to this system since ChartOCR is recent, state-of-the-art, and closest in structure to \sys. In this comparison, we observed that, within the application to python style graphs, the scale and diversity of graphs supported by \sys is greater than that supported by ChartOCR. At a high level, \sys supports more complex plot types such as scatter plots (both single and multi series) or certain multi-series bar charts.
However, from this comparison, we were also able to directly observe that in situations where multiple series significantly overlapped, \sys was more robust and better able to identify the individual series. Specifically, we found that for multi-series line graphs, the raw score performance of \sys was twice as accurate as ChartOCR. We believe that this advantange of \sys is due to its relation extraction approach. By using this approach, \sys can more easily segment out series and focus on high level relations as opposed to other systems that focus on individual data point extraction and thus are easily confused by noisy or overlapping series.

%% file: RelatedWork.tex
\section{Background and Related Work}
First, we describe the basic problem statement and how existing approaches work.

\begin{figure}
    \centering
    \includegraphics[width=0.55\columnwidth]{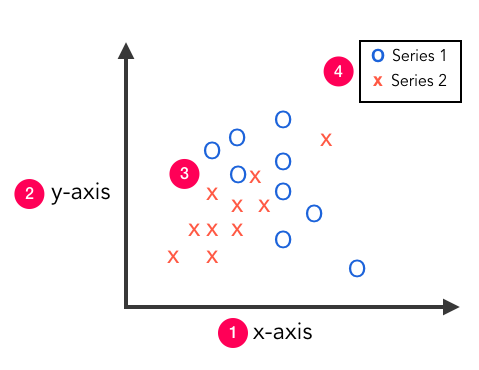}
    \caption{An illustration of terms that we use to describe a visualization with numbered components. (1) X-Axis label, (2) Y-Axis label, (3) the plot data, (4) legend. }
    \label{fig:anatomy} \vspace{-1em}
\end{figure}

\subsection{Problem Statement}
\sys accepts 2-dimensional plots that visualize data in the X-Y plane (Figure \ref{fig:anatomy}). 
These plots illustrate how multiple series (subsets of data) compare on those coordinates.
\sys considers scatter plots, line graphs, and bar charts, each with up-to 3 different series.
The goal of \sys is to extract visual relations from these charts.

\begin{definition}[Visual Relation]
Let $V$ be a visualization and $S \subseteq V$ be a series. A visual relation is the association of the series with a binary function $\theta$ of the series:
\[
S \rightarrow \theta(S)
\]
\end{definition}

This definition is a bit abstract so let us consider an example as in Figure \ref{fig:rels}. We could define a property called \textsf{isIncreasing()} which, for a given series, measures the Spearman correlation coefficient between each mark's x and y coordinates and tests to see if it is above $0.4$. Each series would have a true and false value for \textsf{isIncreasing()}. Suppose that we have a family of such visual relations functions $\Theta = \{\theta_1,...,\theta_N\}$. The visual relation extraction problem is to:

\begin{definition}[Visual Relation Extraction]
Let $S_1,...,S_k$ be a set of series in a chart and $\{\theta_1,...,\theta_N\}$ be a set of visual relation functions. The visual relation extraction problem is to identify all the visual relations in the chart:
\[\{S_i \rightarrow \theta_j(S_i)\}_{1<i\le k,1<j\le N}\]
\end{definition}

This task is straight-forward when the data that generated the chart is available. However, the main challenge in the visual relation extraction problem is to infer these relationships purely from an image of the visualization. This system will have to identify all of the chart series and use visual cues to infer the results of the visual relation functions. For example, \textsf{isIncreasing()} looks different in a line graph than it does in a bar chart.

\begin{table*}\footnotesize
\begin{tabular}{ccccc} 
 \toprule
 System & Scatter Graphs & Line Graphs & Bar Graphs & Pie Charts\\
 \midrule
 \textbf{VizExtract} & \boldcheckmark & \boldcheckmark & \boldcheckmark & $\boldsymbol{\times}$\\
 Scatteract ~\cite{cliche2017scatteract} & \checkmark& $\times$ & $\times$ & $\times$\\
 \textcolor{blue}{FigureQA ~\cite{kahou2018figureQA}} & $\textcolor{blue}{\times}$ & \textcolor{blue}{\checkmark} & \textcolor{blue}{\checkmark} & \textcolor{blue}{\checkmark}\\
 ChartText ~\cite{balaji2018dcharttext} & $\times$ & $\times$ & \checkmark & \checkmark\\
 Extraction via Deep NN  ~\cite{liu2019dacvsdnn}
 & $\times$ & $\times$ & \checkmark & \checkmark\\
 Viz for Non-Viz ~\cite{choi2019viznonviz} & $\times$ & \checkmark & \checkmark & \checkmark\\
 \textcolor{red}{ChartOCR ~\cite{luo2021chartOCR}} & $\textcolor{red}{\times}$ & \textcolor{red}{\checkmark} & \textcolor{red}{\checkmark} & \textcolor{red}{\checkmark}\\
 REV ~\cite{poco2017reverse} & \checkmark & \checkmark & \checkmark & $\times$ \\
 \bottomrule
\end{tabular}
 \begin{tabular}{cccc} 
 \toprule
 Data Extraction & Text Extraction & Multiple Series Support & Interface\\
 \midrule
 \boldcheckmark & \boldcheckmark & \boldcheckmark & \textbf{Extraction}\\
 \checkmark& $\times$ & $\times$ & Extraction\\
 $\textcolor{blue}{\times}$ & \textcolor{blue}{\checkmark} & \textcolor{blue}{\checkmark} & \textcolor{blue}{QA}\\
 \checkmark & \checkmark & \checkmark* & Extraction\\
 \checkmark & \checkmark & \checkmark & Extraction\\
 \checkmark & \checkmark & \checkmark & Extraction\\
 \textcolor{red}{\checkmark} & \textcolor{red}{\checkmark} & \textcolor{red}{\checkmark} & \textcolor{red}{Extraction}\\
 $\times$ & \checkmark & N/A & Extraction\\
 \bottomrule
\end{tabular}
\caption{We survey the features of existing visual information extraction systems. Existing systems have varying scopes and interfaces which makes an apples-to-apples comparison difficult. We experimentally compare to ChartOCR (in red) in our experiments and evaluate our results on the FigureQA dataset (blue). An * indicates that the system can perform the tasks in some capacity, but not to the full extent that others can.}\label{table:comp}
\end{table*}

\subsection{Related Work and Current Scope}
To the best of our knowledge, no other system focuses on relation extraction in the way that \sys does. 
However, there do exist a variety of systems that algorithmically extract data from plots ~\citep{luo2021chartOCR, cliche2017scatteract, poco2017reverse, kahou2018figureQA, liu2019dacvsdnn, choi2019viznonviz, balaji2018dcharttext, kataria2008automatic, al2016automatic}. 
Recent and comparable systems can be categorized based on the types of graphs they support, their ability to extract data and text from the graph, their ability to process multi series graphs, and their overall intended function (extraction or question answering). We compare the capabilities of \sys and other prominent systems in Table \ref{table:comp}.

From this table we can see that the \sys system is the most general/versatile and provides coverage for all areas except pie charts which we chose not to support since correlations are not well defined in this context ~\footnote{For other relations where pie charts are relevant, pie charts could be supported by adding them to the training data}. Scatteract is a system that fully automates the process of extracting data points from images of scatter plots, but does not support other plot types. Furthermore, it does not extract series or axis labels (needed for the relation extraction problem). ChartText is a system that first uses a CNN to classify each image as a specific plot type (either pie chart or various types of bar charts) and then from there extracts data with a network specific to the determined plot type. It also has (limited) multi-series support. Since it only supports those types of graphs, we exclude an experimental comparison. There is an entire line of work on Reverse-Engineer Visualizations (REV). These systems solely extract label, axis, and legend text from images of graphs and do not extract any data points or relations. Technically, they work on ``multi-series'' graphs, but since they are only extracting text, adding multiple series to a graph has no effect as these systems do not differentiate legend text into series associations, but instead lump all such text together.

Thus, while there has been a lot of work on this problem, the aforementioned systems are not quite comparable to \sys. Closer to our contribution, ChartOCR is a system that extracts exact graph data from images and supports multiple different types of graphs. Similarly to \sys, this system uses a combination of an algorithmic approach and deep learning to do so. This system is powerful in that is supports many different graph types, however, each graph type requires a separate network in contrast to \sys which uses a single network for all graph types. We picked ChartOCR as a baseline system and ran it on our generated graphs. We consider ChartOCR to be a strong, state-of-the-art baseline to compare to since: (1) it was published in the last year, and (2) it is supports more graph types, and performs better than many standard data point extraction systems.
Additionally, since ChartOCR combines a rule-based, algorithmic approach with deep neural networks, its structure is most similar to \sys.

Beyond ChartOCR, we also identified another line of work in the machine learning community that studies this problem. FigureQA is another system that analyzes images of graphs. However, instead of focusing on extracting specific information from the graphs, FigureQA attempts to answer common questions about these graphs, for example, ``does line X intercept line Y?''. While this system does not perform the same function as VizExtract, it often ends up extracting information about the underlying dataset. However, due to differences in the interface (exhaustive extractions vs. question answering), it was unclear how to make a comparison of performance. Instead, we use the FigureQA dataset as a test case to evaluate our model.

This work is, in spirit, similar to other information extraction projects like Web Tables~\cite{cafarella2008webtables} that study how to extract relational data from unstructured sources (for a full survey of web extraction refer to~\cite{laender2002brief}). Table extraction is a key problem in data lake management~\cite{nargesian2019data} and we think that extracting relations from charts is the next frontier of this area.

\subsection{Example Application}
Our motivating application for \sys is fact-checking~\cite{karagiannis2020scrutinizer}. Plots and charts can be designed to mislead by scaling/cutting axes or linking unrelated data series~\cite{streeb2019visualize}. \sys gives fact-checkers a tool to understand what relations are being described in a given plot and can corroborate those ``visual relations'' with master data. 
Consider the following examples.
(1) Suppose, that an analyst is diagnosing the failure of a recent marketing campaign and plots several KPIs on a line graph of sales to illustrate correlated factors.
\sys can determine which series are being compared to each other, and these series can be linked back to their data source to determine if that comparison is meaningful (e.g., they might have wildly different data qualities).
(2) A chart in a presentation to executives might be scaled in a way to increase the apparent correlation between two variables (e.g., cutting an axis). 
\sys can determine the correlation between the two variables, and that correlation can be tested against the data source to avoid this situation.
We envision that \sys would be integrated with a knowledge base like system with a queryable interface.

%% file: SystemArchitecture.tex
\section{System Architecture}
Next, we briefly summarize the workflow in \sys and all of the major components. 

\begin{figure}
    \centering
    \includegraphics[width=\columnwidth]{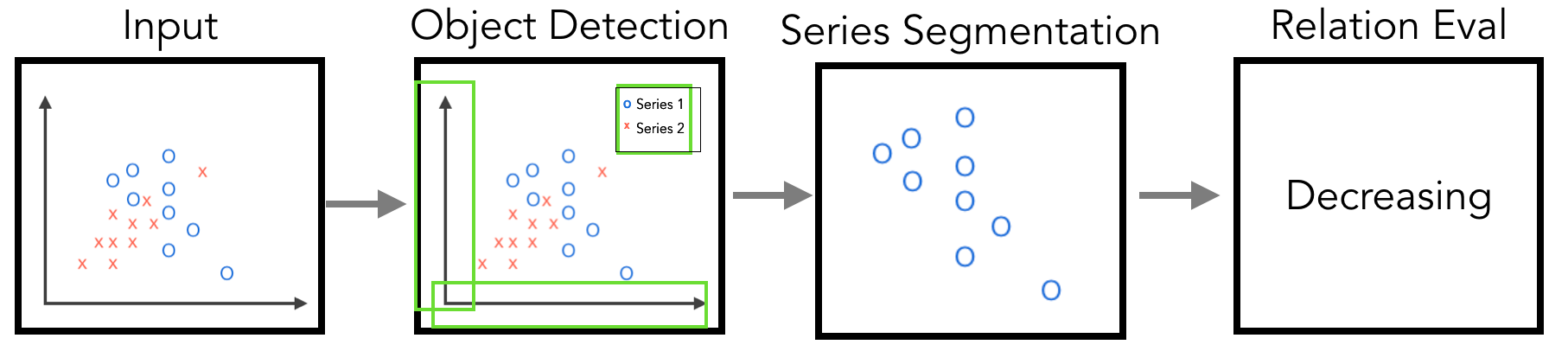}
    \caption{An overview of the main components of \sys}
    \label{fig:overview}
\end{figure}

\subsection{Dataset Generation and Model Training}
The first step is to generate training data and then to train the models for the desired type of relation we want to extract.
The user provides a set of visual relation functions $\theta_1,...,\theta_N$.
The system generates a number of charts with random attributes (plot types, colors, styles, and number of series) of these charts. 
For each of these attributes, the choices of either fixed or randomly varying values are both permitted by the system and can be chosen according to the preferences desired for the given experiment.
All charts are generated with the \texttt{matplotlib} Python library.

\sys is not explicitly restricted to \texttt{matplotlib} plots.
We find that with appropriate randomization the models trained on \texttt{matplotlib} plots generalize to a wide variety of chart styles beyond those generated by this library.
Furthermore, if the user has a very particular plotting library, she can generate a training dataset specific to that library and fine-tune the models described in this work.

Each graph will consist of one to three series with random titles, axis labels, and legends (series names).
Since the ground truth data is known, it is straightforward to determine the true value of $\theta$ and the true values for all of the textual attributes.
The system generates a large training dataset of synthetic charts. Two models are then trained: \textsf{facet\_detector} and \textsf{relation\_extractor}.
\textsf{facet\_detector} model is an object detection model that, given an image, outputs bounding boxes around the key visualization facets: the location of the legend, the location of the axes, and any textual data.
The \textsf{relation\_extractor} is a model that evaluates each of the $\theta_1,...,\theta_N$ on each series in the chart image.

In our experiments, we focus on ``increasing'', ``decreasing'', and ``neutral'' as our desired relations. We determine these relations using the Spearman correlation formula which returns a value ranging from -1 to 1. For all types of graphs, we used a threshold of 0.4, such that data with correlations above 0.4 are classified as positive correlation (increasing), those with correlations below -0.4 are classified as negative correlation (decreasing), and those with correlations between -0.4 and 0.4 are classified as neutral correlation. 

\subsection{Model Inference}
Figure \ref{fig:overview} illustrates the main computer vision pipeline in \sys. \sys takes an input chart as an image file.
It first runs the \textsf{facet\_extractor} to identify all of the core components, such as the legend, the axes labels, and the plot area.
\sys then applies the pre-processing algorithm to the input image. This will convert the original input into a high contrast image where all non-background pixels have high saturation.
Then \sys determines the number of series in the image by running the k-selection algorithm. From here, \sys uses k-means to identify the colors of the background and of each series. For each series color, \sys segments out all pixels in the image that are within a certain range of that color and saves those pixels as a separate image representing the given series. Then \sys runs each new image of a single series through the classifier and for each receives a classification of ``positive", ``negative", or ``neutral".

In addition to the series classification, \sys also performs OCR to determine the text in the image. \sys feeds the raw input image into the object detection network to identify relevant text regions in the image. Then OCR is applied to these regions to extract the text. The identified text is then algorithmically matched to graph elements and \sys returns all of the graph/axis information such as title, axis labels, and legend information. This legend information (which is initially a map from text to \textit{colors}) is then joined with the series correlations resulting in a map from text to \textit{correlations}.

Finally the series classifications and graph text are combined to create the final output of \sys consisting of the general plot text such as title, x axis label and y axis label, and for each series both the text of the series and its corresponding correlation. The user can then use these correlations having extracted the relationship between the variables in their plots.

%% file: Prototype.tex
\section{VizExtract: Visual Processing Pipeline}
Next, we describe in more detail the core components of \sys.

\subsection{Chart Pre-Processing}
In order to apply our model to the diverse types of graphs found in real world situations, we needed to devise a method to remove noise and excess information from images of graphs before we classify them with our model. There are two main types of pixels we want to remove: randomly introduced noise and excessively blurry edges. The random noise is often introduced from processes such as jpg/jpeg compression. While the synthetically generated graphs had sharp edges, many graphs in the wild did not. 

To de-noise the charts, we implemented a novel algorithm to pre-process each image and remove such problematic pixels. This algorithm takes advantage of the fact that these graphs are made to be clearly readable by humans. High saturation colors stand out and are easily distinguishable to the human eye and thus graphs are often created with high saturation series (with different hues) such that these series can be easily differentiated by humans interpreting the graph. This pre-processing algorithm supports graphs with dark backgrounds as well since any grayscale background (including black) gets converted to white and then can be processed normally.

The algorithm works as follows:
\begin{algorithm}
    \caption{$Saturation$ $Threshold$}\label{satThresh}
    \begin{algorithmic}[1]
        \State $thresh \gets $ predetermined saturation threshold constant
        \State convert pixels from RGB space to HSV space
        \For{each pixel, $p$, in the image}
            \If{$p$'s saturation $> thresh$}
                \State $p$'s Saturation $\gets 255$
                \State $p$'s Value $\gets 255$
            \Else
                \State $p$'s Saturation $\gets 0$
                \State $p$'s Value $\gets 0$
            \EndIf
        \EndFor
    \end{algorithmic}
\end{algorithm}

\subsection{Facet Detection}
After pre-processing, the next step is to detect the different facets within the chart image. 

\subsubsection{Object Detection Design}
We train an object detection network based on the Single-Shot Detector architecture to recognize key areas of a graph such as plot space, legend, axis labels, and title.
These detections each return a bounding box of pixels around the relevant areas of the chart.
The bounding boxes can be used to anchor downstream processing.
To train our object detection network, we generate data using the same techniques for creating graphs as described earlier in this paper. When generating the graphs, we generated them in a way that ground truth labels (the true bounding boxes) were simultaneously generated. 
We found that the model was sensitive to variation in input data and as a result, model accuracy was reliant on providing diverse training data. To address this, we created label strings consisting of 3-12 randomly selected upper and lowercase letters. Additionally, we varied the location of the legend, axis labels, and title. 

\subsubsection{Textual Post-Processing}
Once the relevant sections of the image have been identified by the object detection process, we utilize the open source Tesseract optical character recognition software to detect text in these relevant sections [cite]. While Tesseract handles most of the work, there are a few processing steps we preform that greatly increase accuracy. We start by cropping out each relevant section of the image. For the $y$ axis label we rotate its cropped image $90$\textdegree{} clockwise before passing it to Tesseract since in Matplotlib the $y$ axis label is rotated $90$\textdegree{} counter-clockwise by default. Tesseract does support functionality where it can detect and support rotated text, however, we found that on this data, manual rotation led to higher accuracy. \par

\begin{figure}
    \centering
    \includegraphics[width=\columnwidth]{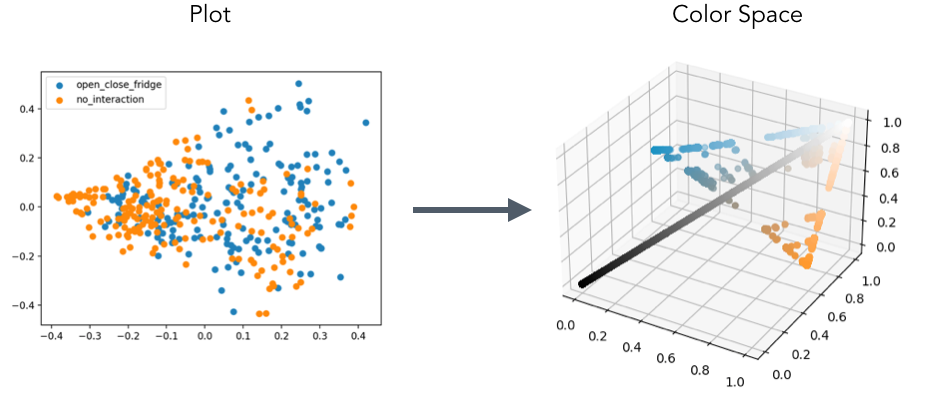}
    \caption{Transforming the image into color space identifies key clusters of similar looking pixels that can be used to identify the marks that denote different series.}
    \label{fig:cs}
\end{figure}

\subsection{Series Extraction}
The series extraction process takes images of graphs and segments out each series for individual analysis. This allows one to independently determine the visual relations of multiple series contained in a single graphical image. This process can be broken down into four main steps: series count determination, identifying series marks, constructing segmentation ranges, and legend series assignment~\footnote{Series count determination and constructing segmentation ranges are considered less central and are described in the appendix.}.

\subsubsection{Series Identification}
The first step is to identify the distinct series in a chart. To do so, we first transform the image into ``color space''. Every pixel in an image can be thought of as a 3D point $(r,g,b)$.
One can think of an image as a collection of these points, i.e., a set of 3D points.
We can visualize this color space with an image like Figure \ref{fig:cs}. 
This shows the different colors present in a particular chart.

Different mark types and colors will result in clusterings in this space. For example, Figure \ref{fig:cs} shows a cluster of orange and a cluster of blue pixels.  
We make use of k-means clustering to identify the relevant colors in the image of the graph to segment out each series. 
We run a k-means clustering algorithm on the RGB values of the pixels in the image of the graph with $k$ equal to one more than the number of series to additionally account for the color of the background ~\footnote{The appendix describes how $k$ is chosen using the combination of a standard ``elbow'' method and normalized error thresholds.}. From the colors that the k-means algorithm returns, we then eliminate the color that corresponds to the background (the color of the axes will already have been removed by pre-processing).

\begin{figure}
\centering
\includegraphics[width=.3\columnwidth]{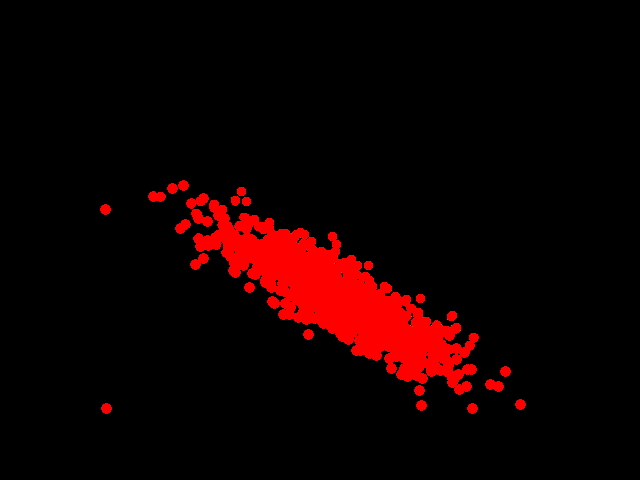}
\includegraphics[width=.3\columnwidth]{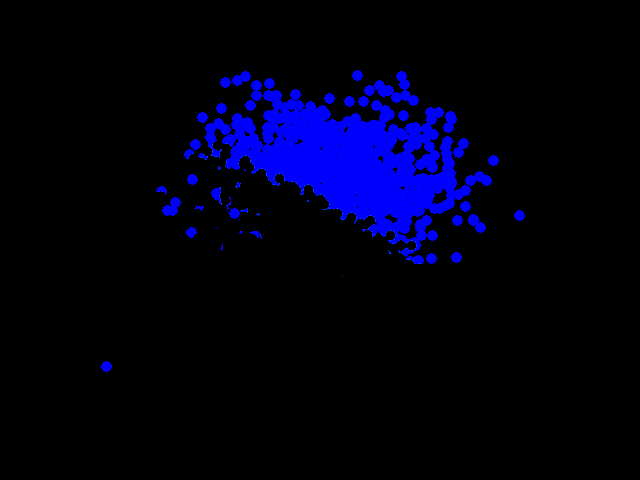}
\includegraphics[width=.3\columnwidth]{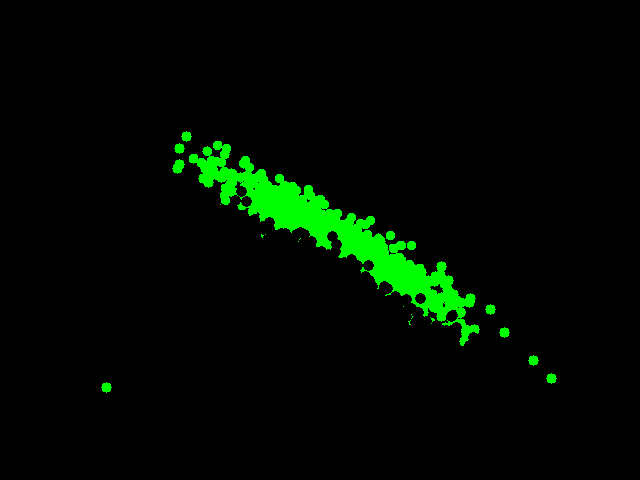}
  \caption{Example of series segmentation.}
  \label{fig:seg}
\end{figure}

In order to differentiate the background from series marks, we simply eliminate the color group that contains the largest percentage of pixels since in all of these types of graphs, there are significantly more pixels making up the background than in the series themselves or in the axes and labels. This method works successfully on the \texttt{matplotlib} graphs created for training data that have solid backgrounds. However, for graphs with multicolor backgrounds, this method could fail.
Once the background color has been eliminated, we return the remaining colors that correspond to the series in the graph.
These colors are turned in to segmentation masks that can extract each data series on its own (Figure \ref{fig:seg}).

\subsubsection{Legend Series Assignment}
Once we have found each series, we must match each text element in the legend with the series. We loop over every text element detected in the legend from the object detection model. For each, we calculate the centroid of this text element and find the nearest "mark" pixel to this element (one that resides in a color cluster corresponding to a mark).
This process is heuristic but we find that it is reasonably robust across multiple different chart styles. The main failure mode is that when the bounding box for the legend is too large, the segmentation might detect a few pixels from points in the data itself. The points are often just as saturated as the key and, if not accounted for, can massively skew the associations. By choosing the median pixel location we make sure that as long as there are more pixels in the series key mark than the confounding pixels, we will still get the correct location.

\subsection{Relation Extraction}
Finally, given the segmented series, \sys performs visual relation extraction. We train a convolutional neural network model to identify each relation extraction function. This network uses a total of nine layers. The first six alternate between a layer of convolution and a layer of max pooling. There are then three fully connected layers including one flattening layer and one output layer. In order to train this model, we first generate training graph images as described in section 3.1. We label each image with ground truth during dataset generation and divide the images into two groups with half for training and half for validation.\par

%% file: Evaluation.tex
\section{Results}
Next, we report the performance of \sys on a variety of real and synthetic datasets.

\subsection{Accuracy Metrics}
\sys  is a system that extracts many pieces of information from a plot. 
First, we evaluate how well \sys can extract different facets of the visualization such as axes, legends, and series. We measure the accuracy of each of these facets individually and report a composite score.  

\subsubsection{Axis Detection}
First, we calculate the fraction of charts where the x-axis label and y-axis label are correctly predicted (within an edit distance threshold).~\footnote{The average length of labels in our experiments $22.5$ characters, and we chose $5$ as an appropriate edit distance threshold.} This results in our first accuracy metric called ``axis'' accuracy that measures correctly extracted x and y axes:
\[A_{axis} = \frac{\text{\# within edit threshold}}{n}\]
The axis accuracy might be low for multiple reasons: (1) the OCR might fail at recognizing the characters, (2) the axes might be improperly detected in the plot, or (3) the axes labels are confused with other facets. 
We do not differentiate these failures unless otherwise noted.

\subsubsection{Series Detection}
Similarly, we defined another metric for detecting and identifying series from the legends. 
Like with the axes, there are many steps including identifying the series from the legend, determining its label text, and assigning it its appropriate relation.
We wrap all three functions into a single metric called A$_{Series}$ for series accuracy.
A chart series is determined to be correct if it is accurately detected in the legend, the corresponding series label is assigned to it~\footnote{We tolerate an OCR error up to an edit distance of $2$ for series labels.}, and the correct relation is determined. A chart itself containing $n$ series receives $\frac{1}{n}$ points for every correct series. For $n$ charts with scores $c_1...c_n$, we define a series accuracy:
\[A_{Series} = \frac{\Sigma_{i=1}^n c_i}{n}\]

\subsubsection{Composite Accuracy}
Finally, once we have calculated both axis and series accuracy, we combine them to get a total accuracy score:
\[A_{Total} = \frac{A_{axis}+A_{series}} {2}\]
Of course, this metric is reductive and accounts for many different  sources of errors. 
We will present experiments with the appropriate ablation studies to illustrate what factors contribute to the accuracy. 

\subsubsection{Series Detection without OCR}
While the metrics defined above are helpful for evaluating the entirety of the functionality of \sys, there are many situations, such as when comparing to other systems, when OCR accuracy is not as important or even practical. As such we define yet another metric, Series$_{NoOCR}$, to measure the series accuracy that does not depend on OCR. This metric works exactly the same way as A$_{Series}$, except instead of comparing the series text label as extracted by OCR to the ground truth text label, it compares the color determined for the series by \sys to the ground truth color label.

\subsection{Evaluation on Generated \texttt{matplotlib} Charts}
In this experiment, we tested the end to end accuracy of our model when both trained and evaluated on charts generated using matplotlib that mirrored those used in the training set (in format but not in content). The results are presented in Table \ref{table:end2end} over $100$ unseen graphs.

\begin{table}[t]
 \begin{tabular}{c c} 
 \toprule
 Metric & Accuracy\\
 \midrule
 Axis & 86.0\%\\
 Series & 83.3\%\\
 Total & 84.5\% \\ [1ex]
 Series$_{No OCR}$ & 87.5\%\\
 \bottomrule
\end{tabular}
\caption{The end to end accuracy metrics for \sys. The ``Accuracy'' column refers to the accuracy metric for graphs generated\label{table:end2end}}\vspace{-1em}
\end{table}

These results show that our end-to-end pipeline can extract plot information from unseen synthetic graphs with reasonably high accuracy. Specifically, it can extract axis information such as labels and titles with $86\%$ accuracy implying that static text extraction is very doable. From manual inspection of the graphs on which the axis extraction fails, we can see that there are often small typos and sometimes these typos exceed the $5$ character buffer specified in the accuracy metric. This error is not especially concerning as these typos are often small (e.g. an ``l" gets switched with an ``I" or ``1") and the text can still be easily interpreted by a human reader. However, in a fully automated application, such typos might be problematic. Another axis error type that we see is that the object detection network misses an axis object entirely (perhaps returning only the title and x label bounding boxes and missing the y label box entirely). This error less concerning since a better trained object detection network should solve this problem. With less limited training examples and more computing power, a more accurate and robust object detection network could be trained and this issue could likely be avoided.\par

Similarly, the relatively low series accuracy can be largely blamed on the OCR portion. Since the series accuracy without OCR is significantly higher at $87.5\%$, the fall off in accuracy logically can be attributed to the OCR portion of the pipeline. From manual inspection, we can observe that the vast majority of these errors occur when the the object detection network fails to recognize the legend object, thus preventing our model from being able to associate each series that has been classified with the appropriate text. Furthermore, of the $12.5\%$ error in the Series$_{N0_OCR}$ accuracy, we can see from manual inspection that the majority of these errors occur on series that are close (within $0.15$) to the $0.4$ boundary that determines the ground truth correlation. In practice, humans themselves would even have difficulty discerning these correlations. If we rerun this experiment excluding such graphs, we get a Series$_{NoOCR}$ accuracy of $94.3\%$.

\subsection{Comparison to ChartOCR}
Next, we compare \sys to ChartOCR, a recently published plot data information extraction system. Since ChartOCR does not support scatter plots, we only evaluated it on line and bar graphs. To compare the two systems, we ran both VizExtract and ChartOCR on the generated plots described before on the relation extraction task.~\footnote{To evaluate, we use both a raw comparison (do the correlation type counts exactly match the ground truth counts for each graph) as well as series accuracy metric described in section 5.1.2 without OCR.} For the line graphs, we generated 100 line graphs with 1-3 series as specified in the dataset generation section. Additionally, since ChartOCR only reports the turning points on line graphs, we interpolated to find more points on those lines and then computed the correlation of those points. This experiment reported the results shown in table \ref{table:chartocr} with graph type ``line''.

\begin{table}[t]
 \begin{tabular}{c c c c}
 \toprule
 System & Graph Type & Raw Score & Series Accuracy\\
 \midrule
 \textbf{VizExtract} & \textbf{Line} & \textbf{90.0}\% & \textbf{95.3}\%\\
 ChartOCR & Line & 43.3\% & 73.3\%\\
 \textbf{VizExtract} & \textbf{Single Series Bar} & \textbf{96.7} \% & \textbf{96.7}\%\\
 ChartOCR & Single Series Bar & 100\% & 100\%\\
 \bottomrule
 \end{tabular}
 \caption{Accuracy metrics comparing \sys to ChartOCR when evaluated on synthetic line and single series bar.} \label{table:chartocr}
 \vspace{-1em}
\end{table}

\begin{figure}
\centering
\includegraphics[width=0.4\columnwidth]{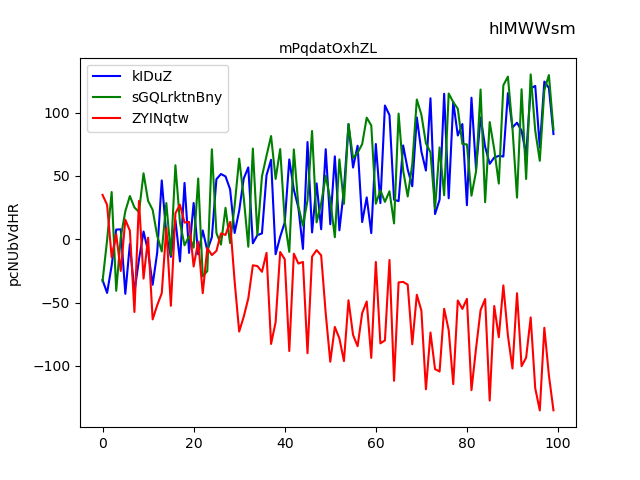}
\includegraphics[width=0.4\columnwidth]{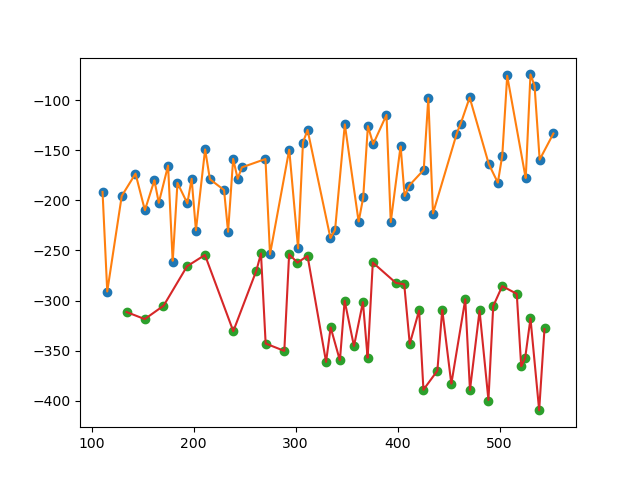}
  \caption{On the left is an example graph with significant overlap. On the right are the points returned by ChartOCR (and the corresponding interpolated lines) demonstrating how ChartOCR struggles with these overlapping series.}
  \label{figure:chartocr}
\end{figure}

The situations where ChartOCR failed was primarily when the series in the graph had significant overlap (see figure \ref{figure:chartocr}). In these situations, it is hard for ChartOCR to pick out exact turning points or even distinguish between two series at all. However, the relation extraction approach of \sys is able to succeed in these scenarios in particular due to its method of series segmentation.


On our generated bar graphs, ChartOCR supports a different format of multi-series bar graphs and thus will only work on our single series bar graphs. As such, we generated 100 single series bar graphs as specified in the dataset generation section. This results for this experiment are reported in table \ref{table:chartocr} with graph type ``single series bar''.

ChartOCR outperforms \sys here, but only slightly as both accuracy metrics are very accurate. 
Furthermore, the high accuracy of both systems is likely due to the fact that single series bar graphs, the one type of bar graph supported by both of these systems, are a simpler problem.

\subsection{Generalization to Alternative Datasets}
Next, we evaluate how \sys performs on alternative datasets of plots. In these experiments, we evaluate the transfer setting where the model is trained on the synthetic data and the system is applied to unseen plot types. We actually found that \sys performed surprisingly well on completely unanticipated data (assuming the broad category of plots were seen in the training dataset). 
We did find that a large number of failures could be attributed to the OCR system so we ignore OCR errors in detecting axes labels and series labels.
Development of better OCR tools is outside the scope of this project and we focus on proper facet detection.

\subsubsection{Graphs Collected from the Internet}
In this experiment, we evaluate our model on graphs collected from the internet. The criteria for choosing these graphs were that they displayed correlations between variables for 1-3 series and that they were either line, scatter, or bar graphs. Given these loose requirements, some of these images are very ``difficult'' to understand due to overlaps in series or odd scaling. Thus, we divided these graphs into two subcategories: easy and hard. These categories were constructed from visual inspection since the generating data is not available. Graphs are by default considered easy and are placed into the hard category if they meet any of the following criteria: the graph contains more than $3$ series, there is a large overlap of series making it hard to see data points, or the correlation of a series in the graph is very close to the boundary and could almost be classified either way. Of our $50$ total graphs, $39$ were classified as easy and $11$ were classified as hard.
The accuracy for our model evaluated on these $50$ graphs gathered from the internet is shown in table \ref{table:googleImages}.

\begin{table}
 \begin{tabular}{c c} 
 \toprule
 Metric & Accuracy \\
 \midrule
 Easy Series$_{NoOCR}$ & 81.2\%\\
 Hard Series$_{NoOCR}$ & 43.2\%\\
 Total Series$_{NoOCR}$ & 72.8\%\\
 \bottomrule
\end{tabular}
\caption{Accuracy metrics for \sys when evaluated on graphs collected from the internet. Graphs are ``easy'' by default and are labeled ``hard'' if they meet certain criteria that make them difficult to classify.} \label{table:googleImages}
\end{table}

From these results we can see that the accuracy on ``easy'' graphs is significantly higher than the accuracy on ``hard'' graphs as expected. The overall Series$_M$ accuracy is lower than the Series$_M$ accuracy for the synthetic graphs, but that is to be expected since they are closer to the actual data that this model was trained on. Through manual inspection, we can observe that the errors on this dataset mostly fall into three categories. The first is that sometimes axis pixels are non-grayscale and thus bypass the pre-processing filter and taint the classification model inputs. The second is when certain series have very specific positions within the graph that are not modeled in the training data. In these cases, the classifier often understandably outputs nonsense since it has never encountered this before. Excluding graphs that fall into at least one the first two categories, these images get classified almost as accurately as the synthetic graphs and similar to the synthetic graphs, the remaining errors often arise when series are very close to the classification boundaries.

\subsubsection{Evaluation on the FigureQA Dataset}
In this experiment, we evaluate the model trained on synthetic data on charts from Microsoft's FigureQA dataset. This dataset was designed for question answering about charts, however, its labels contain sufficient information to calculate the correlation of variables for each series present in the graph. As such, we can use this large and diverse dataset as another metric on which to evaluate our model. However, since this dataset is designed for other purposes, not all of its graphs are of the form that our model is built to handle. As such, we only consider ``line'' and ``dotted line'' graphs with $1-3$ series whose series colors are at least as far apart as our color selection mask ranges.
Additionally, we chose to include only graphs in which all of their series each filled up at least $60\%$ of the height of the graph since oddly scaled charts deviate too much from our training data and the graph types we intended to support.
In the FigureQA dataset, $1129$ charts meet this criteria.
The result of this experiment is shown under ``FigureQA'' in table \ref{table:alldatasets}. This result shows that even with a drastically different dataset such as FigureQA, this same model is still able to achieve good accuracy.

\begin{table}
 \begin{tabular}{c c} 
 \toprule
 Dataset & Series$_{NoOCR}$\\
 \midrule
 Synthetic & 87.5\% \\
 From The Internet & 72.8\% \\
 FigureQA & 84.7\% \\
 \bottomrule
 \end{tabular}
 \caption{Accuracy metrics for all datasets.} \label{table:alldatasets}
\end{table}

Now that we have the Series$_{NoOCR}$ accuracy for all three data sets, we can compare them. From table \ref{table:alldatasets}, we see that there is a slight drop off as we move from synthetic graphs to more general testing data such as the plots from the internet images and the FigureQA dataset. However, this is expected and the fact that these subsequent scores do not have a large dropoff implies that \sys can generalize to unseen graphs and plotting styles.

\subsection{Model Transfer and Fine Tuning}
Model transfer refers to how well a model performs on new data types that it has not seen in training. The importance model transfer is that one would want to avoid having to retrain one's model on each new type of graph. Thus, if our model transfers well to a new data type, we can immediately apply it without any need for additional training. We conduct three experiments to test our model transfer: (1) on synthetic graphs that include diverse plotting styles (distinct from the training data plotting styles), (2) on graphs collected from the internet, and (3) on FigureQA graphs.

\subsubsection{Experiment Setup}
For all three experiments, we first train a model, model-$1$, on a dataset of $100$ synthetic graphs in the default \texttt{matplotlib} style as outlined in section $3.1$. For experiment 1, we then test model-$1$ using a new dataset of $100$ synthetic graphs using a diverse assortment of plotting styles instead of the default \texttt{matplotlib} style. For experiment 2, we instead test model-$1$ using the same ``from the internet'' dataset referenced in section $5.4.1$ of $50$ graphs collected from the internet. For experiment 3, we instead test model-$1$ using the first 100 graphs from the subset of the FigureQA dataset referenced in section $5.4.2$.

For our controls for each experiment, we train a new model with the same type of data being used to test the model in that experiment. For experiment 1, we both train and test a new model, model-$2$, on two new datasets of $100$ synthetic graphs, both of which use a diverse assortment of plotting styles instead of the default \texttt{matplotlib} style. For experiment 2, we both train and test a new model, model-$2$, on the ``from the internet'' dataset (train on 50\%, test on 50\%). For experiment 3, we both train and test a new model, model-$2$, on the subset of the FigureQA dataset referenced in section $5.4.2$ (train on graphs 101-200, test on graphs 201-300). The results of all three experiments are shown in table \ref{table:transfer}.

\begin{table}
 \begin{tabular}{c c c}
 \toprule
 Transfer Experiment & Same Dataset & Synthetic Transfer\\
 \midrule
 Diverse Plotting Styles & 85.4\% & 84.6\% \\
 From The Internet & 82.8\% & 71.8\% \\
 FigureQA & 93.3\% & 84.6\% \\
 \bottomrule
 \end{tabular}
 \caption{Model Transfer accuracy scores for same dataset control and transfer from synthetic intervention} \label{table:transfer}
\end{table}

\subsubsection{Analysis}
The results from all three of these experiments show that \sys is relatively transferable. As expected, training on the same type of data on which you are testing is preferable and thus the ``same dataset'' scores report higher accuracy. If our model was not transferable at all, we might expect to see synthetic transfer accuracy scores near $33\%$ (accuracy of randomly guessing one of the three possible correlations), however, in all cases and especially in the ``Diverse Plotting Styles'' experiment, the transfer score is high and has relatively small drop off from the control.

%% file: Conclusion.tex
\section{Conclusion}
In this paper we proposed \sys, a system that performs relation extraction on graph images by combining computer vision algorithms and deep learning methods.
We have also introduced both a procedure to generate synthetic datasets that can be used as a benchmark for future systems as well as novel evaluation metrics to assess the accuracy of systems attempting relation extraction.
\sys is able to attain high accuracy within the scope of a diverse dataset of synthetic graphs generated in python and is able to support a broad range of graph types.
Our experiments on a state-of-the-art system, ChartOCR, show that \sys's unique approach of relation extraction as opposed to exact data point extraction increases robustness to noise and series overlap.
Lastly, our model is relatively simple and can be trained with hundreds of graphs as opposed to the hundreds of thousands that many other similar systems require. Thus, if any additional training is necessary for novel graph types can be done relatively efficiently.

%% file: app-segment.tex
\section{K Selection Algorithm}


In order to segment each series in our image of a graph, we make use of k-means clustering. We use this technique to identify the relevant colors in the image of the graph that correspond to series that we want to segment out. However, in order to do so, we need to determine an appropriate value for $k$. In the current context, this determination depends on the number of series in the image since we will want a cluster center for each series color as well as an extra one for the background color. A common approach is to use a technique called the Kneedle algorithm ~\cite{satopaa2011kneedle}. However, this technique was not designed for our specific k-means selection in the context of colors and thus what is considered a "knee" by this algorithm is not always the best $k$ that captures the number of series plus one extra cluster center for the background color. As such, this method did not give great accuracy for our specific problem. To correct this, we added an algorithm on top of the kneedle algorithm to further customize the output to the needs of our problem. The algorithm works as follows. We  first normalized the error rates for each $k$. Then we found the $k$ given by the kneedle algorithm. Next, we check if the normalized error of this $k$ choice is below some threshold. If so, we return this $k$ given by the kneedle algorithm. If not, we return the $k$ given by the kneedle algorithm $+ 1$. This ensures that in the situations where the slope of the $k$ vs error graph is less steep and the kneedle algorithm would normally return too small a $k$, the normalized error would fall above the threshold and thus this algorithm would instead return the true best $k$.

\begin{algorithm}
    \caption{$k$ $Selection$. Here $Kneedle()$ refers to applying the ``Knee''-finding Kneedle algorithm specified in ~\citep{satopaa2011kneedle}}\label{kSelection}
    \begin{algorithmic}[1]
        \Require $arr$ (a pixel array representing the image)
        \State $normalizedErrors = []$
        \For{$k$ in $range$($1$-$8$)}
            \State perform $k$-means clustering on $arr$ with $k$ clusters
            \State append the normalized error of the clustering to the $normalizedErrors$ list
        \EndFor
        \State $predictedK \gets Kneedle(normalizedErrors)$
        \If{$predictedK.normalizedError < 0.985$}
            \State $predictedK += 1$
        \EndIf
        \State \Return $predictedK$
    \end{algorithmic}
\end{algorithm}

\section{Constructing Color Ranges}
Another important step of segmentation is deciding which pixels should go with which mean. We refer to this process as finding the appropriate segmentation ranges of the identified colors since we are going to create a color range such that any pixels within that range are segmented out. These ranges are required since the k-means will return means that are similar, but not exactly the same as the exact color of every pixel that makes up a given series. Since hue, saturation, value or HSV color representations are more intuitive to spatially reason about than RGB colors, we convert the RGB colors returned by the color identification step into HSV colors. Then for each HSV color, we create a lower and upper bound around the given mean such that even if the color of the pixels representing the series is slightly different than the associated color returned by k-means it will still be picked up.\par

The lower and upper bounds for HSV ranges given an HSV triple (labelled: [h,s,v]) are created as follows. Lower [H,S,V] = [min(0,h-10), min(0,s-10), min(0,v-40)]. Upper [H,S,V] = [max(180,h+10), max (255,s+10), max(255,v+40)]. Additionally, since hue is essentially the angle in degrees on a color wheel, and since the color red is centered at 0º, we need to implement some system to account for colors wrapping around the color wheel in either direction. We implement this through adding an additional range corresponding to the same series. It works such that if the range of a hue would lead to wrap around, it is capped off by the min and max operations as seen above, but then a second range is added that uses the same S and V values, but encapsulates the wrap around. These two ranges can then be combined later on into a single mask to segmentation. (Note: the API we are using, openCV, measures H from 0º to 180º instead of the usual 360º and S and V both range from 0-255 as opposed to the usual 0-100).\par

In terms of how to choose the size of the ranges for H,S, and V, through experimentation with our dataset we decided on $\pm 10$ for hue, $\pm 25$ for saturation, and $\pm 40$ for value. One thing to note with these ranges is that they have been chosen specifically for the colors used in the data we trained on which were red, green, and blue. Since these colors are evenly spaced out, relatively larger ranges do not run the risk of conflating the colors of multiple distinct series. However, if these methods were to be used on colors that were much closer in HSV space, large ranges would have a much higher risk of covering the colors of multiple distinct series.\par